\documentclass{article}
\usepackage{spconf,amsmath,amssymb,graphicx,cite,balance,url,booktabs}
\title{Audio-Visual Scene-Aware Dialog and Reasoning using\\
Audio-Visual Transformers with Joint Student-Teacher Learning}
\if 0
\name{
  Ankit P. Shah${^\dagger}{^\S}$, 
  Shijie Geng${^\ddag}$, 
  Peng Gao${^\star}$,
  \em Anoop Cherian${^\dagger}$,
  Takaaki Hori${^\dagger}$,
  Tim K. Marks${^\dagger}$,
  Jonathan Le Roux${^\dagger}$,
  Chiori Hori${^\dagger}$
}
\else
\makeatletter
\def\@name{\emph{Ankit P. Shah${^\dagger}{^\S}$}, 
  \emph{Shijie Geng${^\ddag}$}, 
  \emph{Peng Gao${^\star}$}, \\
  \emph{Anoop Cherian${^\dagger}$},
  \emph{Takaaki Hori${^\dagger}$},
  \emph{Tim K. Marks${^\dagger}$},
  \emph{Jonathan Le Roux${^\dagger}$},
  \emph{Chiori Hori${^\dagger}$} \\
}
\makeatother
\fi

\address{
${^\dagger}$Mitsubishi Electric Research Laboratories (MERL)\\
${^\S}$Carnegie Mellon University
\hspace{10 pt}
${^\ddag}$Rutgers University
\hspace{10 pt}
${^\star}$The Chinese University of Hong Kong
}

\begin{document}
\ninept
\maketitle
\begin{abstract}
%To build a dialog system that can enable an agent to discuss audio-visual scenes with humans, 
In previous work, 
we have proposed the Audio-Visual Scene-Aware Dialog (AVSD) task, collected an AVSD dataset, developed AVSD technologies, and
%, which combined question answering (QA) and dialog response generation on daily life videos annotated with 10-turn question-answering dialogues.
hosted an AVSD challenge track at both the 7th and 8th Dialog System Technology Challenges (DSTC7, DSTC8). 
%While prior approaches designed to tackle this task have shown the need for multimodal fusion to improve response quality, 
In these challenges, the best-performing systems relied heavily on human-generated descriptions of the video content, which were available in the datasets but would be unavailable in real-world applications. 
To promote further advancements for real-world applications, we proposed a third AVSD challenge, at DSTC10, with two modifications: 1) the human-created description is unavailable at inference time, and 2) systems must demonstrate temporal reasoning by finding evidence from the video to support each answer.
This paper introduces the new task that includes temporal reasoning and our new extension of the AVSD dataset for DSTC10, for which we collected human-generated temporal reasoning data. We also introduce a baseline system built using an AV-transformer, which we released along with the new dataset. Finally, this paper introduces a new system that extends our baseline system with attentional multimodal fusion, joint student-teacher learning (JSTL), and model combination techniques, achieving state-of-the-art performances on the AVSD datasets for DSTC7, DSTC8, and DSTC10. We also propose two temporal reasoning methods for AVSD: one attention-based, and one based on a time-domain region proposal network. 
\end{abstract}

\section{Introduction}
\vskip -2mm
To encourage development of dialog system technologies that enable an agent to discuss audio-visual scenes with humans, we held two challenges on audio-visual Scene-Aware Dialog (AVSD) at DSTC7 and DSTC8 \cite{DSTC7_Overview, DSTC8_Overview} using a dataset we collected based on the videos from the Charades dataset \cite{sigurdsson2016hollywood}. 
The AVSD task we defined and dataset we prepared were the first attempt to promote the combination of audio-visual question-answering systems and conversation systems into a single framework \cite{hori_2019_ICASSP, Alamri_2019_CVPR}. 
This task that we proposed is to generate a system response to a query, where the query is part of a multi-turn dialog about a video. Challenge participants used the video, its associated audio, and the dialog text to train end-to-end deep learning models to produce the answers. In addition, the systems had access to human-created video captions. 
%, however, unlike previous challenges, participants cannot use the human-created video captions. 
%Furthermore, temporal reasoning for answers needs to be provided for each generated answer. 
%Using an end-to-end framework, 
The AVSD task can be seen as an extension to video data of both the {\em visual question answering} (VQA) task~\cite{VQA, balanced_binary_vqa, balanced_vqa_v2, tapaswi2016movieqa}, in which the goal is
%The goal of VQA is 
to generate answers to questions about a scene in a static image, and the {\em visual dialog} task~\cite{DBLP:journals/corr/DasKGSYMPB16},
in which an AI agent holds a meaningful dialog with humans about a static image using natural, conversational language \cite{visdial_rl}. Another progenitor to AVSD is the task of \emph{video description} (text summarization of videos), which~\cite{Hori_2017_ICCV} addressed utilizing multimodal attention mechanisms, which selectively attend to different input modalities (feature types) such as spatiotemporal motion features and audio features, in addition to temporal attention.
Combining video description technologies like these with end-to-end dialog systems enables scene-aware dialog systems that make use of multimodal information, such as audio and visual features. In a more recent work, spatio-temporal reasoning has been shown to improve performance on AVSD tasks~\cite{geng2021dynamic}.
Recently, Transformer-based AVSD systems outperform LSTM-based ones \cite{Le_2019, li2021bridging}.

The task setup for AVSD in DSTC7--8 allowed participants to use human-created video captions to help generate answers for the dialog questions, and systems that used these human-generated captions significantly outperformed systems that did not. However, since such human-created descriptions are not available in real-world applications of an AVSD system, in practice a system needs to learn to produce the answers without the captions.
There are two other design difficulties that such text-based descriptions introduce that may skew the evaluation: (i) some descriptions already include parts of the answers that are used in the evaluations, making audio-visual inference redundant, and (ii) language models trained using a simple (and limited) QA dataset may generate answers using frequently-occurring text patterns in the training data, without needing to use audio-visual cues (e.g., Q: How many people are in the scene? A: Two people). 
 %These observations are empirically supported by the results: without providing human-generated descriptions, the best performing model achieves only 0.387 in BLEU score, which is a relative reduction of 12\% from its score when using human descriptions. 
The results from AVSD in DSTC7--8 suggest there is still an opportunity to design better audio-visual reasoning methods to approach the performance achieved when using manual video descriptions, but without using these descriptions at test time.
Furthermore, real systems should ideally be able to show the evidence supporting their generated answers, by pointing to the relevant segments of the video. To encourage progress towards this end, we propose a third AVSD challenge in DSTC10\footnote{\url{https://github.com/dialogtekgeek/AVSD-DSTC10_Official}}.
% for the video-based scene-aware dialog track.
% confirming their answer

In this paper, we introduce the DSTC10-AVSD challenge task, the goals of which are: 1) answer generation without human-created captions at inference time, and 2) temporal reasoning (providing evidence) for the generated answers.
Furthermore, we develop an AVSD baseline system using an AV-transformer \cite{iashin2020abetter}. In addition, we propose a novel system that extends this AV-transformer using attentional multimodal fusion~\cite{Hori_2017_ICCV}, joint student-teacher learning (JSTL)~\cite{hori2019joint}, and model combination techniques. We also propose two temporal reasoning methods for AVSD: one attention-based, and one based on a region proposal network (RPN). 
Results show that our extended AV-transformer achieves state-of-the-art on DSTC 7, 8, and 10 when combined with our LSTM-based AVSD system~\cite{hori2019joint}.%by combining it with our LSTM-based AVSD system from~\cite{hori2019joint}.

%%%%%%%%%%%%%%%%%%%%%%%%%%%%%%%%%%%%%%%%%%%%%%%%%%%%%%%
\section{Audio-Visual Scene-Aware Dialog data set}
%%%%%%%%%%%%%%%%%%%%%%%%%%%%%%%%%%%%%%%%%%%%%%%%%%%%%%%
\vskip -2mm
%\label{sec:video-features}
We base the new Audio-Visual Scene-Aware Dialog (AVSD) task for DSTC10 on the AVSD dataset from DSTC7--8~\cite{DSTC7_Overview, DSTC8_Overview}. For the AVSD data, we collected text-based dialogs on short videos from the popular Charades dataset~\cite{sigurdsson2016hollywood},
which consists of untrimmed and multi-action videos (each video also has an audio track) and comes with human-generated descriptions of the scene. 
In our video scene-aware dialog case, two parties, dubbed {\em questioner} and {\em answerer}, have a dialog about events in the provided video.
The job of the answerer, who has already watched the video, is to answer questions asked by the questioner\cite{Alamri_2019_CVPR}.
Table \ref{tab:data} shows the size of the data used for DSTC10. 
For this year's challenge (DSTC10), we collected additional data for temporal reasoning, in which humans watched the videos and read the dialogues, then identified segments of the video containing evidence to support a given answer. 

\begin{table}[t]
\centering
\caption{Audio-Visual Scene-aware Dialog data set for DSTC10.}
\label{tab:data}
\begin{tabular}{ll|ccc}
\toprule
& 
& training 
& validation 
& test \\
\midrule
\#dialogs 
& 
& 7,659
& 1,787 
& 1,804  \\
\#turns   
& 
& 153,180 
& 35,740 
& 28,406  
\\
\#words   
& 
&  1,450,754 
&  339,006
&  272,606
\\
\bottomrule
\end{tabular}
\vskip -4mm
\end{table}
%%-------------------------

%%%%%%%%%%%%%%%%%%%%%%%%%%%%%%%%%%%%%%%%%%%%%%%
\section{Baseline Model}
\begin{figure}[t]
    %\vskip -2mm
	\centering
	\includegraphics[width=8.5cm]{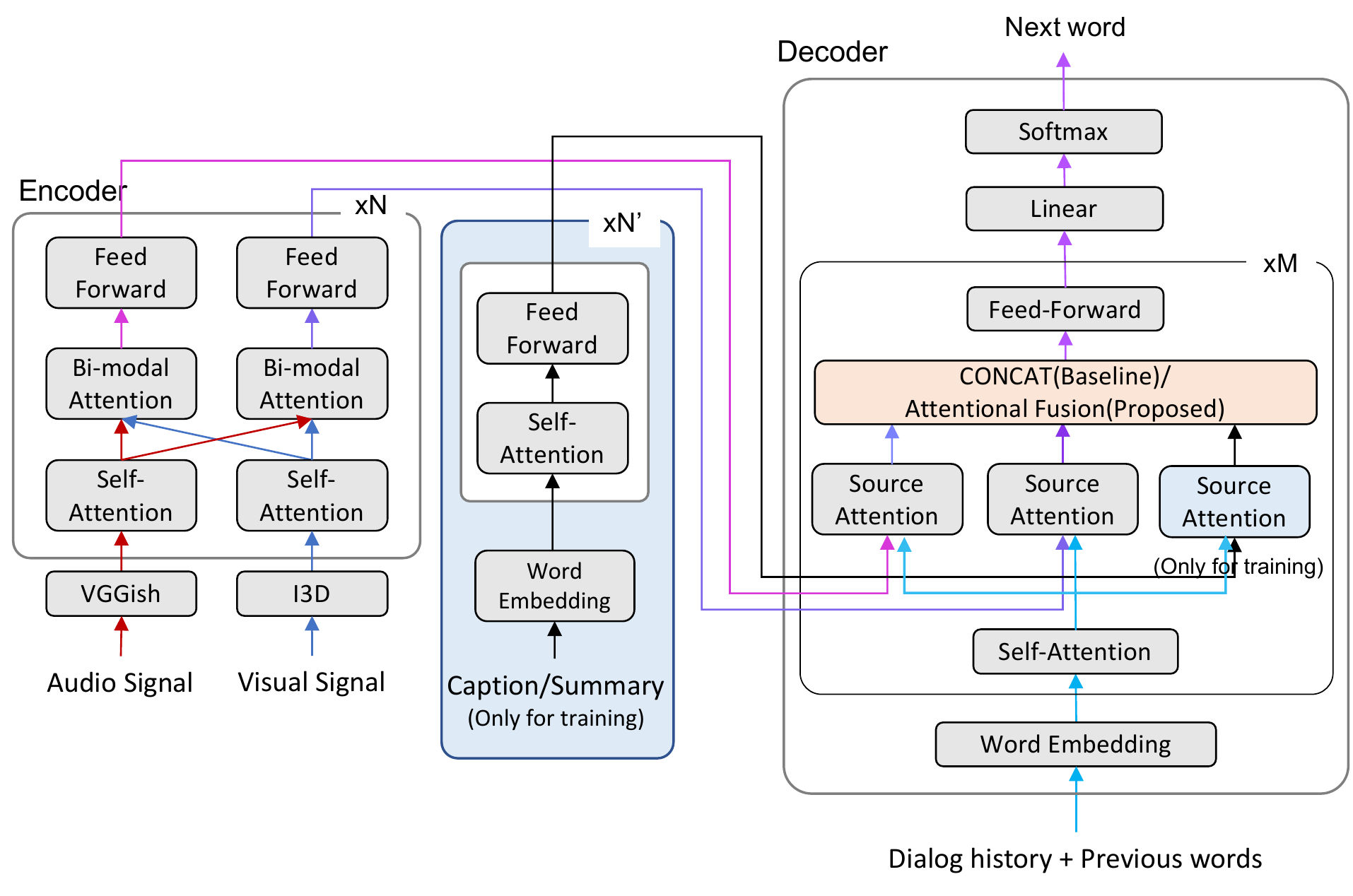}\vspace{-.3cm}
	\caption{Baseline and extended AV-transformer. Our extended system adds the JSTL modules (blue and orange boxes) to the baseline.}
	\label{fig:baseline+proposed}
	\vskip -2mm
\end{figure}

%%%%%%%%%%%%%%%%%%%%%%%%%%%%%%%%%%%%%%%%%%%%%%%
\vskip -2mm
Our DSTC10-AVSD baseline model is an AV-transformer architecture \cite{iashin2020abetter}, shown in Fig.~\ref{fig:baseline+proposed}. %built to investigate better response generation methods with help of video description. 
The system employs a transformer-based encoder-decoder, including a bimodal attention mechanism \cite{NN_MT@ICLR2015,NIPS2015_5847} that lets it learn interdependencies between audio and visual features.

Given a video stream, the audio-visual encoder extracts VGGish~\cite{hershey2017VGGish}  and I3D~\cite{carreira2017quo} features from the audio and video tracks, respectively, 
%where the frame rate may be different on each track.
%The sequences of audio and visual features from a starting point to the current time 
%which are fed to the encoder, and converted to hidden vector sequences through 
and encodes these using
self-attention, bimodal attention, and feed-forward layers. Typically, this encoder block is repeated $N$ times, e.g., $N\!\geq\!6$. 
%The final encoded representation is obtained via the $N$th encoder block.
More formally, let $X^A$ and $X^V$ denote audio and visual signals. First, the feature extraction module extracts VGGish and I3D feature vector sequences from the input signals:
\begin{align}
A^0=\mathrm{VGGish}(X^A), ~~V^0=\mathrm{I3D}(X^V).
\end{align}
The $n$th encoder block computes hidden vector sequences as:
\begin{align}
    \bar{A}^n &= A^{n-1}+\mathrm{MHA}(A^{n-1}, A^{n-1}, A^{n-1}),\label{eq:enc_self1}\\
    \bar{V}^n &= V^{n-1}+\mathrm{MHA}(V^{n-1}, V^{n-1}, V^{n-1}),\label{eq:enc_self2}\\
    \tilde{A}^n &= \bar{A}^{n}+\mathrm{MHA}(\bar{A}^{n}, \bar{V}^{n}, \bar{V}^{n}),\label{eq:enc_bm1}\\
    \tilde{V}^n &= \bar{V}^{n}+\mathrm{MHA}(\bar{V}^{n}, \bar{A}^{n}, \bar{A}^{n}),\label{eq:enc_bm2}\\
    A^n&=\tilde{A}^{n}+\mathrm{FFN}(\tilde{A}^{n}), \\
    V^n&=\tilde{V}^{n}+\mathrm{FFN}(\tilde{V}^{n}),
\end{align}
where $\mathrm{MHA}$ and $\mathrm{FFN}$ denote multi-head attention and feed-forward network, respectively.
Layer normalization~\cite{ba2016layer} is applied before every $\mathrm{MHA}$ and $\mathrm{FFN}$ layer, but it is omitted from the equations for simplicity.
$\mathrm{MHA}$ takes three arguments: query, key, and value vector sequences~\cite{vaswani2017attention}. 
The self-attention layer extracts temporal dependency within each modality, where the arguments for $\mathrm{MHA}$ are all the same, i.e., $A^{n-1}$ or $V^{n-1}$, as in \eqref{eq:enc_self1} and \eqref{eq:enc_self2}.
The bimodal attention layers further extract cross-modal dependency between audio and visual features, taking the keys and values from the other modality as in \eqref{eq:enc_bm1} and \eqref{eq:enc_bm2}.
After that, the feed-forward layers are applied in a point-wise manner.
The encoded representations for audio and visual features are obtained as $A^N$ and $V^N$.

%--- Fig.1 at Page 3--------------------------------------------
%---------------------------------------------------------------

The decoder receives the encoder outputs and the dialog history until the current question, and 
starts generating
%iteratively predicts the next word to generate 
the answer sentence from the beginning token (\texttt{<sos>}) placed at the end of the last question.
At each iteration step, it receives the preceding word sequence and predicts the next word by applying $M$ decoder blocks and a prediction network.%, where each word is assumed to be converted to a word embedding vector.
In each decoder block, the encoded audio-visual features are combined with each word using the bimodal attention layers.
%{\allowdisplaybreaks
Let $Y_i$ be a dialog history plus preceding word sequence $h_1,...,h_L,\texttt{<sos>},y_1,...,y_i$ after $i$ iterations and $Y_i^0$ be a word embedding vector sequence given by $Y_i^0=\mathrm{Embed}(Y_i)$.

Each decoder block has self-attention, bimodal source attention, and feed-forward layers. Computations within the $m$-th block are as follows:
\begin{align}
    \bar{Y}_i^m &=Y_i^{m-1} + \mathrm{MHA}(Y_i^{m-1},Y_i^{m-1},Y_i^{m-1}),\label{eq:dec_self} \\
    \bar{Y}_i^{Am} &=\bar{Y}_i^m + \mathrm{MHA}(\bar{Y}_i^m, A^N, A^N), \label{eq:dec_src1}\\
    \bar{Y}_i^{Vm} &=\bar{Y}_i^m + \mathrm{MHA}(\bar{Y}_i^m, V^N, V^N), \label{eq:dec_src2}\\
    \tilde{Y}_i^m &=\mathrm{Concat}(\bar{Y}_i^{Am}, \bar{Y}_i^{Vm}), \label{eq:dec_ff1}\\
    Y_i^m &= \tilde{Y}_i^m + \mathrm{FFN}(\tilde{Y}_i^m). \label{eq:dec_ff2}
\end{align}
The self-attention layer converts the word vectors to high-level representations considering their temporal dependency in \eqref{eq:dec_self}.
The bimodal source attention layers update the word representations based on the relevance to the encoded multi-modal representations in \eqref{eq:dec_src1} and \eqref{eq:dec_src2}.
A feed-forward layer is then applied to the outputs of the bimodal attention layers in \eqref{eq:dec_ff1} and \eqref{eq:dec_ff2}.
Finally, a linear transform and softmax operation are applied to the output of the $M$-th decoder block to obtain the probability distribution of the next word as
\begin{align}
    P(\mathbf{y}_{i+1}|Y_i,X^A,X^V)=\mathrm{Softmax}(\mathrm{Linear}(Y_i^M)).
\end{align}
At inference time, we can pick the one-best word $\hat{y}_{i+1}$ for $y_{i+1}$ as
\begin{equation}
        \hat{y}_{i+1} = \mathop{\mathrm{argmax}}_{y \in \mathcal{V}} P(\mathbf{y}_{i+1}=y|Y_i,X^A,X^V),
\end{equation}
where $\mathcal{V}$ denotes the vocabulary, and 
%After picking the one-best word $\hat{y}_{i+1}$, 
the answer sentence is extended by adding the selected word to the already generated word sequence as $Y_{i+1}=Y_i,\hat{y}_{i+1}$.
This is a greedy search process that ends if $\hat{y}_{i+1}=\texttt{<eos>}$, which represents an end token.
It is also possible to pick multiple words with highest probabilities and consider multiple candidates for the answer sentence using the beam search.

\section{Extended AV-transformer}
\vskip -2mm
We extend the baseline AV-transformer by applying attentional multimodal fusion \cite{Hori_2017_ICCV} and joint student-teacher learning (JSTL) \cite{hori2019joint}, which have successfully been applied to an LSTM-based AVSD system \cite{hori_2019_ICASSP} but have not previously been applied to transformer-based systems.
In this paper, we propose to extend the AV-transformer with these techniques and test their effectiveness.

Fig.~\ref{fig:baseline+proposed} shows the teacher model of the extended AV-transformer, which has a caption/summary encoder in the encoder and an attentional fusion layer in the decoder. In student-teacher learning, a student model without the caption/summary encoder and its attention module in the decoder is trained using the teacher model output as the target distribution. 

\if 0
%--- Fig.2 at Page 3--------------------------------------------
\begin{figure}[ht]
	\centering
	\includegraphics[width=8.0cm]{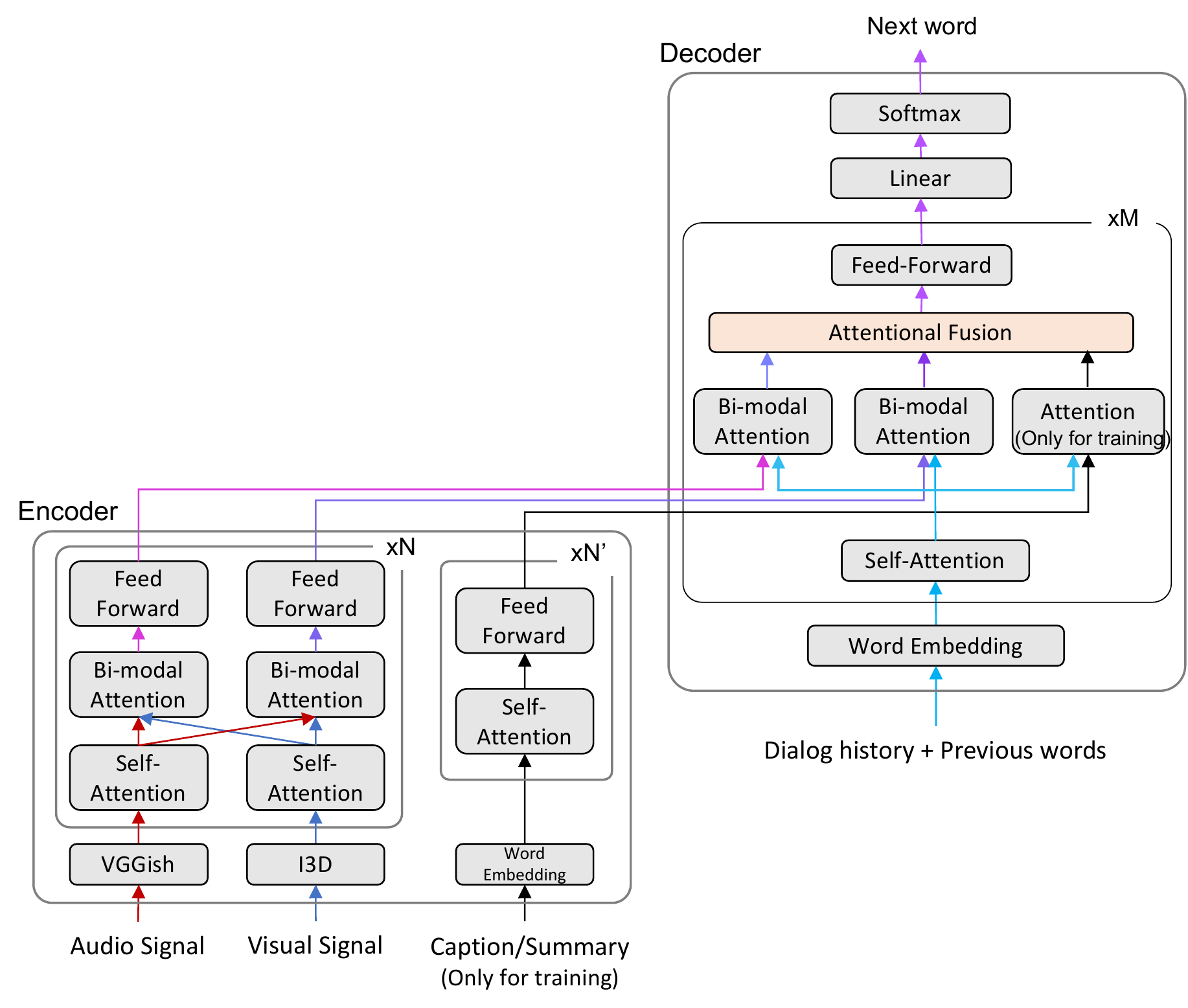}
	\caption{Teacher AV-transformer for student-teacher learning.}
	\label{fig:teacher_model}
\end{figure}
%---------------------------------------------------------------
\fi 
To further improve the performance, we combine the extended AV-transformer with the LSTM-based model trained with student-teacher learning as well, where the two decoder outputs are averaged in the log domain during the beam search.

\subsection{Attentional Multimodal Fusion}
\vskip -1mm
The baseline AV-transformer in Fig. \ref{fig:baseline+proposed} concatenates multi-modal encoder outputs in each decoder block, assuming that the audio and visual features have equal contribution to the next word prediction regardless of the given question and the generated answer. However, prior work has shown that attentional multimodal fusion is effective for LSTM-based systems. In this work, we apply the attentional fusion technique to the AV-transformer.
In the case of Transformer, we can use single-head attention (SHA) in each decoder block as
\begin{equation}
    \tilde{\tilde{Y}}_i^m=\mathrm{SHA}(\bar{Y}_i^m, \tilde{Y}_i^m, \tilde{Y}_i^m),
\end{equation}
where $\tilde{Y}_i^m$ is here a concatenation of $\bar{Y}_i^{Am}$ and $\bar{Y}_i^{Vm}$. If the model has a caption/summary encoder, its output $\bar{Y}_i^{Cm}$ is also concatenated.
In this case, $\tilde{Y}_i^m$ is a $3 \times D$ tensor including three modalities, each of which has a $D$-dimensional vector.
Then, the fused vector $\tilde{\tilde{Y}}_i^m$ is fed to the feed-forward layer. 

\subsection{Student-Teacher Learning}
\vskip -2mm
The goal of student-teacher learning is to obtain a student model that does not make use of the video caption or summary, which is trained to mimic a teacher model that has already been trained using the caption/summary text.
Accordingly, the student model can be used to generate system responses without relying on the caption text, while hopefully achieving similar performance to the teacher model.

\if 0
Following the best system in DSTC7-AVSD track \cite{sanabria2019cmu}, we insert the description text at the beginning of each question.
This means that the same description is always fed to the encoder together with a new question, at every turn of the dialog about the target video clip.
The student network is trained to reduce the cross entropy loss, by using the output of the teacher network as a soft target to make the output distribution of the student model closer to that of the teacher model.
\fi
The student-teacher loss is a cross entropy loss with {\it soft} targets:
\begin{align}
\mathcal{L}_\text{ST}=-\!\sum_{i=1}^{|Y|}\sum_{y \in \mathcal{V}} \!\hat{P}(y|Y_{i-1},X^A\!\!, X^V\!\!, X^C) \log P(y|Y_{i-1}, X^A\!, X^V),
\end{align}
where $\hat{P}(y|Y_{i-1}, X^A, X^V, X^C)$ denotes the probability distribution for the $i$th word obtained by the teacher network. Here, $P(y|Y_{i-1}, X^A, X^V)$ is the posterior distribution from the current student network (which is being trained), which is predicted without the caption text $X^C$.

Following our prior work, we also incorporate a decoder state similarity loss and a cross-entropy loss on the teacher for joint student-teacher learning as
%a joint student-teacher learning as
\begin{align}
\mathcal{L}_\text{JST}=
\mathcal{L}_\text{ST}  + \lambda_c \mathcal{L}_\text{MSE} + \mathcal{L}_\text{CE}^{(T)},
\end{align}
where $\mathcal{L}_\text{MSE}=\sum_{i=1}^{|Y|} \mathrm{MSE}(Y_i^m, \hat{Y}_i^m)$. Here, $\mathrm{MSE}(\cdot,\cdot)$ denotes the mean square error between two vectors, $\lambda_c$ denotes a scaling factor, and $m=M/2$, i.e., we use only one intermediate layer.
We aim here to compensate for missing input features at the decoder state level, so that the student model can hopefully exploits other modalities more actively.
Furthermore, joint student-teacher learning updates not only the student network but also the teacher network. We use the standard cross entropy $\mathcal{L}_\text{CE}^{(T)}$ for a {\it hard} target, only for the teacher network.
Likewise, $\mathcal{L}_\text{ST}$ is used only for the student network, while $\mathcal{L}_\text{MSE}$ is used for both networks.

%%%%%%%%%%%%%%%%%%%%%%%%%%%%%%%%%%%%%%%%%%%%%%%%%%%%
\section{Temporal Reasoning}
%%%%%%%%%%%%%%%%%%%%%%%%%%%%%%%%%%%%%%%%%%%%%%%%%%%%
\label{sec:reasoning}
\vskip -2mm
Temporal reasoning is the task of finding evidence supporting the generated answers, where the evidence corresponds to human-annotated time regions of the video that have been identified as supporting each ground-truth answer.
Human annotators were allowed to choose multiple time regions for each question-answer pair,
but most of the reasons consist of a single region.

\subsection{Attention-based method}
\vskip -1mm
We built a baseline method for temporal reasoning based on attention weights obtained during decoding.
The attention weights are computed to predict each word, where each attention weight corresponds to a certain time frame of input audio/visual features.
Thus, a high weight means that the corresponding time frame is strongly correlated to a word in the generated answer.
Given an attention weight distribution, we can compute mean $\mu$ and standard deviation $\sigma$ of the distribution, and roughly estimate the time region as $\mu \pm \nu \sigma$, where $\nu$ is a hyper parameter. Since we have multiple attention distributions over the word sequence, attention heads, and layers, we use their averaged distribution. 
This method finds only one time region for each answer, and it requires no special training to select time regions.

\subsection{RPN-based time region detection}
\vskip -1mm
We also built a CNN-based temporal reasoning model, which accepts encoder outputs of the AV-transformer and an embedded QA pair to predict temporal regions that support the answer.
The model employs a time-domain region proposal network (RPN) \cite{ren2015faster,iashin2020abetter}, where Conv1D modules with different kernel sizes accept frame-level outputs of the multimodal encoders, each of which is concatenated with the QA pair embedded by the decoder followed by mean pooling. It predicts the center position, the region length, and the confidence score of each region candidate. We pick high-confidence regions from the candidates using a predetermined threshold. 

\section{Experiments}
\vskip -2mm
We evaluate our AV-transformer using the AVSD datasets from DSTC7, DSTC8, and DSTC10. Training and validation sets are common across the three challenges, but the test sets are different.
%%%%%%%%%%%%%%%%%%%%%%%%%%%%%%%%%%%%
\subsection{Conditions}
%%%%%%%%%%%%%%%%%%%%%%%%%%%%%%%%%%%%
\vskip -1mm
We extracted VGGish audio features~\cite{hershey2017VGGish} and I3D video features~\cite{carreira2017quo} from each video clip, where I3D features consisted of sequences fo 2040-dimensional RGB and flow vector, and VGGish features were sequences of 128-dimensional vectors.
The RGB and flow features were concatenated before feeding them to the encoder.

The baseline AV-Transformer has projection layers before encoder blocks, where the audio and visual features are projected to 64 and 128 dimensional vectors, respectively.
The encoder has 2 encoder blocks, in which the audio and visual attention layers have 64 and 128 dimensions, and their feed-forward layers have 256 and 512 dimensions, respectively.
The decoder has 2 decoder blocks, in which 300-dimensional GloVe word vectors \cite{pennington2014glove} are projected to 256-dimensional embedding vectors and fed to 256-dimensional attention layers followed by 1024-dimensional feed-forward layers.
The baseline system employs greedy search to generate the answers.

The quality of the automatically generated sentences was evaluated with objective measures to compare the similarity between the generated sentences and the ground truth sentences. %(see Fig. \ref{fig:task}).
We used the evaluation code for MS COCO caption generation\footnote{\url{https://github.com/tylin/coco-caption}} for objective evaluation of system outputs, which supports automated metrics such as BLEU, METEOR, ROUGE\_L, and CIDEr.

%%%%%%%%%%%%%%%%%%%%%%%%%%%%%%%%%%%%%%%%%%%%%%%%%%%
\subsection{Results and Discussion}
%%%%%%%%%%%%%%%%%%%%%%%%%%%%%%%%%%%%%%%%%%%%%%%%%%%
\vskip -1mm
\setlength\tabcolsep{2pt} % default value: 6pt
%---------------------------------------------------------
\begin{table}[t]
\centering
\caption{Evaluation results on DSTC7-AVSD test set.}
\label{tab:dstc7-result}
\begin{tabular}{l|cccc}
\toprule
Model & BLEU4 & METEOR & ROUGE\_L & CIDEr \\
\midrule
Baseline AV-transformer & 0.296 & 0.214 & 0.485 & 0.771 \\
+ Hyperparam. tuning.   & 0.362 & 0.237 & 0.522 & 0.974 \\
+ Beam search           & 0.380 & 0.239 & 0.530 & 0.998 \\
+ Attentional MM fusion & 0.391 & 0.248 & 0.536 & 1.013 \\
+ JST learning          & 0.401 & 0.256 & 0.549 & 1.051 \\
+ Comb. with LSTM       & \bf{0.406} & \bf{0.262} & \bf{0.554} & \bf{1.079} \\
\midrule
LSTM + JST learning \cite{hori2019joint} & 0.382 & 0.254 & 0.537 & 1.005 \\
DSTC7 best \cite{sanabria2019cmu} w/ cap.* & 0.394 & 0.267 & 0.563 & 1.094 \\
\bottomrule
\multicolumn{5}{r}{*DSTC7 best system does not have results without captions.}
\end{tabular}
%\vskip -5mm
%\end{table}
%-------------------------------------------------------
%\begin{table}[t]
%\centering
\caption{Evaluation results on DSTC8-AVSD test set.}
\label{tab:dstc8-result}
\begin{tabular}{l|cccc}
\toprule
Model & BLEU4 & METEOR & ROUGE\_L & CIDEr \\
\midrule
Baseline AV-transformer & 0.281 & 0.203 & 0.468 & 0.701 \\
Extended AV-transformer & 0.380 & 0.242 & 0.535 & 0.957 \\
+ Comb. with LSTM       & \bf{0.394} & \bf{0.250} & \bf{0.545} & 0.997 \\
\midrule
%LSTM + JST learning \cite{hori2019joint}  & 0.382 & 0.248 & 0.538 & 0.953 \\
DSTC8 best \cite{li2021bridging} w/o cap. & 0.387 & 0.249 & 0.544 & \bf{1.022} \\
\bottomrule
\end{tabular}
\vskip -4mm
\end{table}
%----------------------------------------------------------
Table \ref{tab:dstc7-result} shows the evaluation results on the DSTC7 test set.
To improve the performance from the baseline, we first tuned the hyperparameters using the validation set, where we made the decoder network deeper to 6 blocks and reduced the dimension of the attention layers to 200. 
We shrank the dialog history given to the decoder into just the previous question. In addition, we applied a learning rate control that halves the learning rate of Adam optimizer if the validation loss did not decrease after each training epoch.
With this tuning, we obtained substantial improvement, e.g., $0.296 \rightarrow 0.332$ in BLEU4.
Then, we applied the beam search technique with beam size 5, which further improved the performance.

We extend the AV-transformer by adding attentional multimodal (MM) fusion and joint student-teacher (JST) learning, achieving further performance improvement.
Finally, we combine our AV-transformer with our LSTM-based model from~\cite{hori2019joint}, which also employed attentional MM fusion and JST learning.
When we combine the word posterior probabilities of the two decoders in the log domain, we obtain the best results, which outperform the prior method \cite{hori2019joint} and even achieve competitive performance to the best DSTC7 system that used the caption/summary information.

Table \ref{tab:dstc8-result} shows the evaluation results on the DSTC8 test set.
As in the DSTC7 results, the AV-transformer including all the extensions shows substantial improvements on all the performance metrics.
Furthermore, the table also shows that combination of the AV-transformer and the LSTM model achieves the state-of-the-art performance in BLEU4, METEOR, and ROUGE\_L in comparison with the DSTC8 best system \cite{li2021bridging} based on a large-scale Transformer initialized with GPT-2 \cite{radford2019language}, for the condition in which caption/summary information were not available.

%----------------------------------------------------------
\begin{table}[t]
\centering
\caption{Evaluation results on DSTC10-AVSD test set.}
\label{tab:dstc10-result}
\begin{tabular}{l|cccc}
\toprule
Model & BLEU4 & METEOR & ROUGE\_L & CIDEr \\
\midrule
Baseline AV-transformer & 0.247 & 0.191 & 0.437 & 0.566 \\
Extended AV-transformer & 0.371 & 0.245 & 0.535 & 0.869 \\
+ Comb. with LSTM       & \bf{0.385} & \bf{0.247} & \bf{0.539} & \bf{0.888} \\
\bottomrule
\end{tabular}
%----------------------------------------------------------
%\vskip 2mm
\caption{Evaluation results on temporal reasoning for DSTC10-AVSD test set.}
\label{tab:dstc10-reasoning}
\begin{tabular}{l|cc}
\toprule
Model               & IoU-1 & IoU-2 \\
\midrule
Attention method    & 0.361  & 0.380 \\
Region Proposal Net (RPN) & \bf{0.521} & \bf{0.550} \\
\bottomrule
\end{tabular}
\vskip -4mm
\end{table}
%----------------------------------------------------------
Finally, we evaluated our model with the DSTC10-AVSD test set.
The sentence generation performance is shown in Table \ref{tab:dstc10-result}, and we see improvements similar to the ones in the DSTC7 and DSTC8 results.
We also evaluated the reasoning performance of the attention-based and RPN-based methods introduced in Section~\ref{sec:reasoning}.
The RPN had 3-layer Conv1D modules with 10 different kernel sizes for each modality and 256 dimensions in each internal layer.
Table~\ref{tab:dstc10-reasoning} shows the reasoning performance measured by Intersection over Union (IoU), which indicates the ratio of overlap between the predicted and ground-truth time regions (higher is better).
Since there may be multiple valid reasons for each answer, we designed two IoU measures, where IoU-1 is obtained as an average IoU computed between each ground truth and the predicted region that gives the highest IoU to the ground truth.
IoU-2 is computed by frame-level matching among all predicted and ground-truth regions for each answer, i.e., frames included in both predicted and ground-truth regions are counted as intersections while those included in both or either of them are counted as union.
Table~\ref{tab:dstc10-reasoning} shows that the RPN outperforms the naive attention-based approach, which suggests that model training with ground-truth annotations for temporal reasoning is important for temporal reasoning in the AVSD task \footnote{\url{https://github.com/dialogtekgeek/AVSD-DSTC10_Official/tree/main/baseline}}.

%%%%%%%%%%%%%%%%%%%%%%%%%%%%%%%%%%%%%
\section{Conclusions}
%%%%%%%%%%%%%%%%%%%%%%%%%%%%%%%%%%%%%
\vskip -2mm
In this paper, we introduced the DSTC10-AVSD task and dataset, which promote further advancements into real-world applications of the AVSD, in which human-created descriptions are not available at inference time and where temporal reasoning is required to provide evidence supporting the answers.
We developed an AV-transformer as a baseline system for DSTC10-AVSD. We also proposed extending it with attentional multimodal fusion, joint student-teacher learning, and model combination techniques, achieving state-of-the-art performance. 
Our experiments compared the performance of the baseline system and our extended system with the previous state of the art, testing on the AVSD test sets for DSTC7, DSTC8, and DSTC10. 
We have just released the temporal reasoning dataset and the baseline system for open competition as the AVSD challenge in DSTC10.

\balance 
\bibliographystyle{IEEEtran}
\bibliography{mybib}
\end{document}